\documentclass[a4paper, 10pt, conference]{ieeeconf}      

\IEEEoverridecommandlockouts                              

\usepackage{amsmath} 
\usepackage{algpseudocode}
\usepackage{algorithm}
\usepackage{tikz}
\usepackage{pgfplots}

\usepackage{subcaption}

\usepackage{multicol}
\usepackage{graphicx}
\usepackage{siunitx}
\newcommand{\mbeq}{\overset{!}{=}}

\usepackage{setspace}

\usepackage{pifont}
\newcommand{\cmark}{\ding{51}}%
\newcommand{\xmark}{\ding{53}}%

\title{\LARGE \bf
Towards Courteous Behavior and Trajectory Planning for Automated Driving 
}

\author{Oliver Speidel$^{1}$, Maximilian Graf$^{1}$, Thanh Phan-Huu$^{2}$ and Klaus Dietmayer$^{1}$
\thanks{*This work was supported by Visteon Corporation}
\thanks{$^{1}$O. Speidel, M. Graf and K.Dietmayer are with the Institute of Measurement, Control and Microtechnology, Ulm University, 89081 Ulm, Germany, e-mail: firstname.lastname@uni-ulm.de.}%
\thanks{$^{2}$T. Phan-Huu is with Visteon Corporation, Karlsruhe, Germany, ~~~~~~~~~~~~~~~~ e-mail: tphanhuu@visteon.com}
}
\usepackage[top=1.46in, bottom=0.80in,left=0.77in, right=0.52in ]{geometry}

\newcommand\copyrighttext{%
	\scriptsize \copyright~2019 IEEE. Personal use of this material is permitted. Permission from IEEE must be obtained for all other uses, in any current or future media, including reprinting/republishing this material for advertising or promotional purposes, creating new collective works, for resale or redistribution to servers or lists, or reuse of any copyrighted component of this work in other works.}%
\newcommand\copyrightnotice{%
	\begin{tikzpicture}[remember picture,overlay]
	\node[anchor=south,yshift=10pt,xshift=0.25cm] at (current page.south) {{\parbox{\dimexpr\textwidth-\fboxsep-\fboxrule\relax}{\copyrighttext}}};
	\end{tikzpicture}%
}

\begin{document}

\maketitle
\copyrightnotice
\thispagestyle{empty}
\pagestyle{empty}

\begin{abstract}

Efficient behavior and trajectory planning is one of the major challenges for automated driving. Especially intersection scenarios are very demanding due to their complexity arising from the variety of maneuver possibilities and other traffic participants. A key challenge is to generate behaviors which optimize the comfort and progress of the ego vehicle but at the same time are not too aggressive towards other traffic participants. In order to maintain real time capability for courteous behavior and trajectory planning, an efficient formulation of the optimal control problem and corresponding solving algorithms are required. 
Consequently, a novel planning framework is presented which considers comfort and progress as well as the courtesy of actions in a graph-based behavior planning module. Utilizing the low level trajectory generation, the behavior result can be further optimized for driving comfort while satisfying constraints over the whole planning horizon.
According experiments show the practicability and real time capability of the framework.

\end{abstract}

\section{Introduction}

In the context of automated driving, behavior and trajectory planning are basic requirements as well as major challenges. Thus, huge effort has been made in this area in recent years \cite{Villagra2012,Werling2010,Ziegler2014}.
In this regard, urban scenarios still represent a vast challenge, as particularly efficient and safe behavior has to be generated. 
Especially when driving through intersections, the problem complexity increases dramatically, as there might be multiple merging and crossing lanes including other traffic participants. Thus, various traffic rules and the behavior of other traffic participants have to be taken into account. Traffic light guided intersections can already be handled reliably, as no complex prediction of other traffic participants is necessary \cite{Kunz2015}. In contrast, driving maneuvers without right of way at intersections or on-ramp scenarios are still demanding. For example, aggressive merging behavior may lead to fast progress of the ego vehicle but at the same time induces high costs for other vehicles as they might be forced to decelerate in order to keep a safety gap \cite{Evestedt2016}. Whereas, passive behavior may lead to situations in which the automated vehicle is not able to take a turn because of congested traffic.
A key element is to find a motion plan which is appropriate for other vehicles but at the same time optimizes the progress of the ego vehicle. Recent research even shows that courteous behavior leads to better imitation of human behavior \cite{Sun2018}.
Therefore, a novel framework is presented which is able to generate trajectories optimized for comfort and progress while considering not only constraints, as for example traffic lights, but also costs for other vehicles induced by the ego trajectory. In general, a graph-based behavior planning module yields a rough trajectory which is optimized afterwards by the low level trajectory generation module. 

The main contribution of this paper is two-fold. On the one hand, a novel behavior planning strategy is presented which allows to plan maneuvers under the consideration of costs for other traffic participants on a large horizon ($t_\tau \approx 10s$). On the other hand, a sampling based optimization strategy using septic polynomials and an associated replanning method is shown, which yields comfortable trajectories even in changing environments on the whole planning horizon. Thereby, the whole framework is real time capable.

\section{Related Work}

Behavior and trajectory planning for automated driving has been widely studied.
In unstructered environment a common method is to first search a drivable path and afterwards an according velocity profile is generated \cite{Villagra2012}.
An approved concept for structured environment is to sample quintic polynomials in a Fr\'{e}net frame which allows longitudinal and lateral movement along a given path \cite{Werling2010, Kunz2015}. To do so, target states are determined by a rule-based heuristic depending on the current behavior state. Another approach is to use local, continuous methods \cite{Ziegler2014}. However, these concepts are not well suited for planning complex maneuvers on long horizons while considering traffic rules and other traffic participants, as it is computationally infeasible \cite{Hubmann2016, Ziegler2009}.
To overcome this problem, a novel approach is to combine behavior and trajectory planning with a graph search problem and an underlying low level trajectory optimization \cite{Hubmann2016, Ward2018a}. Thereby, a rough behavior trajectory is extracted which is then used for further optimization. However, heavy sampling of quintic polynomials is necessary in order to generate smooth trajectories. Furthermore, the optimality and safety is not guaranteed on the whole horizon as in low level optimization only the next three seconds are considered.  
In addition, the interactions of other traffic participants are not taken into account \cite{Hubmann2016, Ward2018a}. 
In order to plan under consideration of interactions and uncertainties, a common approach is to find the optimal policy for a Partially Observable Markov Decision Process (POMDP) \cite{Brechtel2014,Hubmann2018,Bai2015}. However, these approaches are either limited in the scenarios which can be handled, lack of real time capabilities or only yield discretized actions without consideration of low level optimization \cite{Brechtel2014,Hubmann2018,Bai2015}. 
Furthermore, there are concepts accounting for costs of other traffic participants. For example, trajectories can be sampled using quintic polynomials and subsequently rated, where the according reaction of other traffic participants are considered with the Intelligent Driver Model (IDM) \cite{Evestedt2016}. However, the sampling heuristics are restricted to a rule-based strategy and therefore, generating trajectories on long horizons is not practicable.

A method for producing courteous behavior is to use a game-theoretic interaction model \cite{Sun2018}. The problem is modeled in a way that all agents try to optimize their own behavior. By predicting and considering the reaction to the ego trajectory, courteous behavior can be generated. It is shown that this courtesy leads to better imitation of human behavior \cite{Sun2018}. 

For this reason, in this work the game-theoretic problem formulation is borrowed and included into a novel planning framework in order to enable courteous driving for practical use in real time.

\section{Problem Statement}

In general, the motion planning problem for automated vehicles consists of traveling towards a defined goal state in an efficient and convenient manner while obeying traffic rules. For the presented approach, it is assumed to have a given route extracted from a high-precision digital map.
A common approach is to formulate the optimal control problem in Fr\'{e}net coordinates and therefore it is possible to reduced the problem to find a longitudinal motion plan along the predefined center line \cite{Werling2010}.
Thereby, existing definitions can be used but have to be adapted \cite{Hubmann2016}. To satisfy vehicle kinematics, acceleration bounds \mbox{$a_\text{e} \in [a_{\text{min}}, a_{\text{max}}]$} and velocity bounds  $v_\text{e} \in [0, v_{\text{max}}]$ are introduced. Where the maximum velocity $v_{\text{max}} = f(\kappa(s), \mathcal{R})$ is restricted by traffic rules $\mathcal{R}$ and the curvature $\kappa$ of the longitudinal position $s$ on the given path.
Additionally, constraints due to traffic lights or other vehicles can be represented as a combination of position and time intervals $\mathbf{c} = [t_{\text{start}}, t_{\text{end}}, s_{\text{start}}, s_{\text{end}}]$. Hereby, $s_{\text{start}}$ defines the start of the spatial constraint on the center line. The end of the restricted zone is described by $s_{\text{end}}$. Accordingly, $t_{\text{start}}$ and $t_{\text{end}}$ describe the time interval in which the defined zone on the center line is forbidden.

The optimal control problem which optimizes the progress and comfort along the center line is given by,

\begin{equation}
\mathbf{u}^{\text{e}*} = \arg\min_{\mathbf{u}^\text{e}} J^\text{e}(\mathbf{X}, \mathbf{u}^\text{e},\kappa(s^\text{e}), \mathbf{C}^\text{e})\,,
\end{equation}

where $J$ is a cost function, $\mathbf{X}$ represents the current state, $\mathbf{u} = [u_k,...,u_\tau]$ is a sequence of actions, $\mathbf{C}$ a set of spatio-temporal constraints and $\text{e}$ indicates the ego vehicle.
In order to enforce courteous behavior, the costs for other vehicles induced by the ego vehicle have to be taken into account.

To model a system with multiple agents, the game-theoretic optimal control problem presented in \cite{Sun2018} is chosen and adapted. Thereby, the interaction model assumes agents optimizing their own behavior.
In order to make the problem computational tractable, a simplification is performed by approximating the behavior of other vehicles by a reactive driver model. Therefore, the action function of other vehicles is approximated by
\begin{eqnarray}
	\nonumber
	\mathbf{u}^{\text{o}*} &=& \arg\min_{\mathbf{u}^\text{o}} J^\text{o}(\mathbf{X}, \mathbf{u}^\text{e}, \mathbf{u}^o, \kappa(s^\text{o}), \mathbf{C}^\text{o}) \\ &\approx& g^\text{o}(\mathbf{X},\mathbf{u}^\text{e},\kappa(s^\text{o}),\mathbf{C}^\text{o})\,,
\end{eqnarray}
where $\text{o}$ indicates other vehicles.
As a result, the generic action function for the ego vehicle can be extended to
\begin{equation}
\mathbf{u}^{\text{e}*} = \arg\min_{\mathbf{u}^\text{e}} J^\text{e}(\mathbf{X}, \mathbf{u}^\text{e},\kappa(s^\text{e}), \mathbf{C}^\text{e}, g^\text{o}(\mathbf{X},\mathbf{u}^\text{e},\kappa(s^\text{o}),\mathbf{C}^\text{o}))\,,
\label{eqn:ocpformulation}
\end{equation}
where $\text{o} \in 1,\dots,m$ and $m$ is the number of other relevant vehicles.
This formulation enables courteous behavior planning. In order to be able to plan complex maneuvers this problem has to be solved on a large horizon ($t_\tau \approx 10s$).
For this reason, the following framework is presented.

\section{Courteous Motion Planning} 

In order to solve the problem formulation shown in Equation \ref{eqn:ocpformulation}, a two staged optimization is used consisting of a graph based high level behavior planning method followed by a low level optimization module using septic polynomials. The general structure is based on \cite{Hubmann2016}.
However, the problem formulation for behavior planning and especially the low level optimization of the trajectories differs drastically from previous work.

\subsection{Behavior Planning} 
The behavior planning module aims to find a rough trajectory $\mathcal{T}_\text{B}$ which consists of discrete states $\mathbf{x}_k^\text{e}$, where $k = 1,...,\tau$, with the temporal spacing $\Delta t$.
In contrast to \cite{Hubmann2016}, the transition model between states or respectively vertices is formulated with constant jerk (CJ) according to

\begin{multline}
	\setlength\arraycolsep{1.5pt}
\underbrace{\begin{bmatrix}
	s_{k+1}^\text{e} \\
	v_{k+1}^\text{e} \\
	a_{k+1}^\text{e}
	\end{bmatrix}}_{\mathbf{x}_{k+1}^\text{e}}
	=
	\begin{bmatrix}
	1 & \Delta t & \frac{1}{2}\Delta t^2\\ 
	0 & 1 & \Delta t\\ 
	0 & 0 & 1\\
	\end{bmatrix}
	\begin{bmatrix}
	s_k^\text{e} \\
	v_k^\text{e} \\
	a_k^\text{e} \\
	\end{bmatrix}
	+
	\begin{bmatrix}
	\frac{1}{6}\Delta t^3\\[1pt]
	\frac{1}{2}\Delta t^2 \\
	\Delta t
	\end{bmatrix}
	\dot{a}_k^\text{e}\,,
	\label{eqn:caTransition}
\end{multline}
where $u^\text{e}_k = \dot{a}_k^\text{e}$. This allows a smooth interpolation between two behavior states with polynomials and therefore serves as a better basis for the low level optimization as shown in \mbox{Section \ref{subsec:trajectoryGeneration}}.
However, the action set \mbox{$\mathcal{A} = \{a^{(1)},\dots, a^{(n)}\}$} is defined by different acceleration values in the succeeding state. This allows faster changes of the acceleration and thereby more flexible maneuvers. Consequently, $\dot{a}_k^\text{e}$ is calculated by \mbox{$\dot{a}_k^\text{e} = (a_{k+1}^\text{e}-a_k^\text{e})/{\Delta t}$}. 
As a reactive prediction model for other traffic participants, exemplary the IDM is used \cite{Treiber2000}.  Accordingly, $g^o(\cdot)$ is given by the IDM in combination with the constant acceleration (CA) transition model. The IDM is defined by,  
\begin{equation}
a_k^\text{o} = a_{\text{IDM}}\bigg(1-\Big(\frac{v_k}{v_{\text{des}}}\Big)^\delta - \underbrace{\Big(\frac{s^*(v_k,\Delta v_k)}{\Delta s_k}\Big)^2}_{I_{\text{IDM},k}}\bigg)\,,
\label{eqn:IDM}
\end{equation}
where $\Delta s$ denotes the distance to the leading vehicle and $I_{\text{IDM}}$ describes the interaction term of the IDM with
\begin{equation}
	s^*(v,\Delta v) = s_0 + vT + \frac{v\Delta v}{2\sqrt{a_{\text{IDM}}b_{\text{comf}}}}\,,
\end{equation}
and $\Delta v = v_{\text{lead}} - v$, where $v_{\text{lead}}$ is the velocity of the leading vehicle.
\begin{figure}
	\centering
	\def\svgwidth{0.7\columnwidth}
	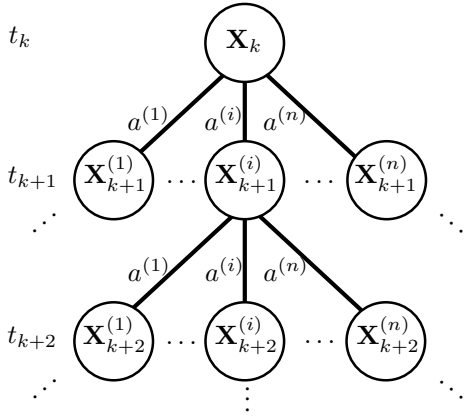
	\caption{Exemplary section of the utilized behavior graph. Vertices represent states connected by edges associated with actions contained in \mbox{$\mathcal{A} = \{a^{(1)},\dots, a^{(n)}\}$}. The graph search is utilized to find the optimal sequence of actions up to the horizon $t_\tau$.}
	\label{fig:graph}
\end{figure}
The transition model is given by
\begin{equation}
\underbrace{\begin{bmatrix}
	s_{k+1}^\text{o} \\
	v_{k+1}^\text{o}
	\end{bmatrix}}_{\mathbf{x}_{k+1}^\text{o}}
=
\begin{bmatrix}
1 & \Delta t \\ 
0 & 1
\end{bmatrix}
\begin{bmatrix}
s_k^\text{o} \\
v_k^\text{o} \\
\end{bmatrix}
+
\begin{bmatrix}
\frac{1}{2}\Delta t^2 \\
\Delta t
\end{bmatrix}
a_k^\text{o}\,.
\end{equation}
As a result, the complete state for a time step $k$ can be described as
\begin{equation}
\mathbf{X}_k = [\mathbf{x}_k^\text{e}, \mathbf{x}_k^{1\dots m}]\,,
\end{equation}
which is a concatenation of the ego vehicle state and the states of the predicted vehicles. A schematic representation of the resulting behavior graph and according states is depicted in Fig. \ref{fig:graph}.
In addition to the possible state transition, the costs for states have to be defined. Therefore, the costs for a given state $\mathbf{X}_k$ are formulated as follows
\begin{equation}
	J_{k}^\text{e} = \omega_\text{f} j_{\text{f},k} + \omega_v j_{v,k} + \omega_\text{jerk} j_{\text{jerk,k}} +\omega_{\text{inter}} j_{\text{inter,k}}\,,
\end{equation}
where $\boldsymbol{\omega} = [\omega_\text{f}, \omega_v, \omega_\text{jerk}, \omega_{\text{inter}}]$ defines different weightings for the single cost terms. Consequently, the overall cost function is
\begin{equation}
	J^\text{e}(\mathbf{X}, \mathbf{u}^\text{e},\kappa(s^\text{e}), \mathbf{C}^\text{e}, g^o(\mathbf{X},\mathbf{u}^\text{e},\kappa(s^\text{o}),\mathbf{C}^\text{o})) = \sum_{k=0}^\tau J_{k}^\text{e}\,.
\end{equation}
By penalizing the jerk
\begin{equation}
	j_{\text{jerk},k} = (\dot{a}_k^\text{e})^2\,,
\end{equation}
costs for varying acceleration values are induced, which leads to more stable and comfortable trajectories.
The costs for the velocity deviation of the desired velocity $v_\text{des}$ are chosen as follows
\begin{equation}
j_{v,k} = \begin{cases}
(v_{k}^\text{e} - v_{\text{des}}(s_{k}^\text{e}))^2 &\mbox{if } v_{k}^\text{e} > v_{\text{des}}(s_{k}^\text{e}) \\
|v_{k}^\text{e} - v_{\text{des}}(s_{k}^\text{e})| &\mbox{if } v_{k}^\text{e} \leq v_{\text{des}}(s_{k}^\text{e}) \end{cases}.
\end{equation}
This cost term ensures to reach the desired velocity $v_\text{des}$ and a soft constraint is utilized in order to stay below $v_\text{des}$.
The costs for following other vehicles is calculated using the interaction term of the IDM
\begin{equation}
	j_{\text{f},k} = I_{\text{IDM},k}\,.
\end{equation}
This allows the dynamic consideration of speed and spatial difference to the leading vehicle. As a result, an appropriate gap to the vehicle in front is kept.
Following the definition of the courtesy term in \cite{Sun2018}, the interaction costs for other traffic participants are formulated as the costs arising due to actions of the ego vehicle. Therefore, induced costs for other vehicles are considered by
\begin{equation}
j_{\text{inter,k}} = \sum_o |a_{\text{norm},k}^\text{o}-a_{\text{inter},k}^\text{o}|\,,
\end{equation}
where $a_{\text{norm},k}^\text{o}$ is the predicted acceleration for any other vehicle if the ego vehicle is without influence. In contrast, $a_{\text{inter},k}^\text{o}$ is the predicted acceleration if the ego vehicle interacts with the other vehicle. In this work, this generic concept is exemplary modeled for two different interaction types at an intersection where the other vehicle has right of way. The scenarios are also depicted in Fig. \ref{fig:crossingMerging}:

\begin{figure}
	\centering
	\def\svgwidth{\columnwidth}
\begingroup%
  \makeatletter%
  \providecommand\color[2][]{%
    \errmessage{(Inkscape) Color is used for the text in Inkscape, but the package 'color.sty' is not loaded}%
    \renewcommand\color[2][]{}%
  }%
  \providecommand\transparent[1]{%
    \errmessage{(Inkscape) Transparency is used (non-zero) for the text in Inkscape, but the package 'transparent.sty' is not loaded}%
    \renewcommand\transparent[1]{}%
  }%
  \providecommand\rotatebox[2]{#2}%
  \newcommand*\fsize{\dimexpr\f@size pt\relax}%
  \newcommand*\lineheight[1]{\fontsize{\fsize}{#1\fsize}\selectfont}%
  \ifx\svgwidth\undefined%
    \setlength{\unitlength}{1673.65541175bp}%
    \ifx\svgscale\undefined%
      \relax%
    \else%
      \setlength{\unitlength}{\unitlength * \real{\svgscale}}%
    \fi%
  \else%
    \setlength{\unitlength}{\svgwidth}%
  \fi%
  \global\let\svgwidth\undefined%
  \global\let\svgscale\undefined%
  \makeatother%
  \begin{picture}(1,0.50138109)%
    \lineheight{1}%
    \setlength\tabcolsep{0pt}%
    \put(0,0){\includegraphics[width=\unitlength,page=1]{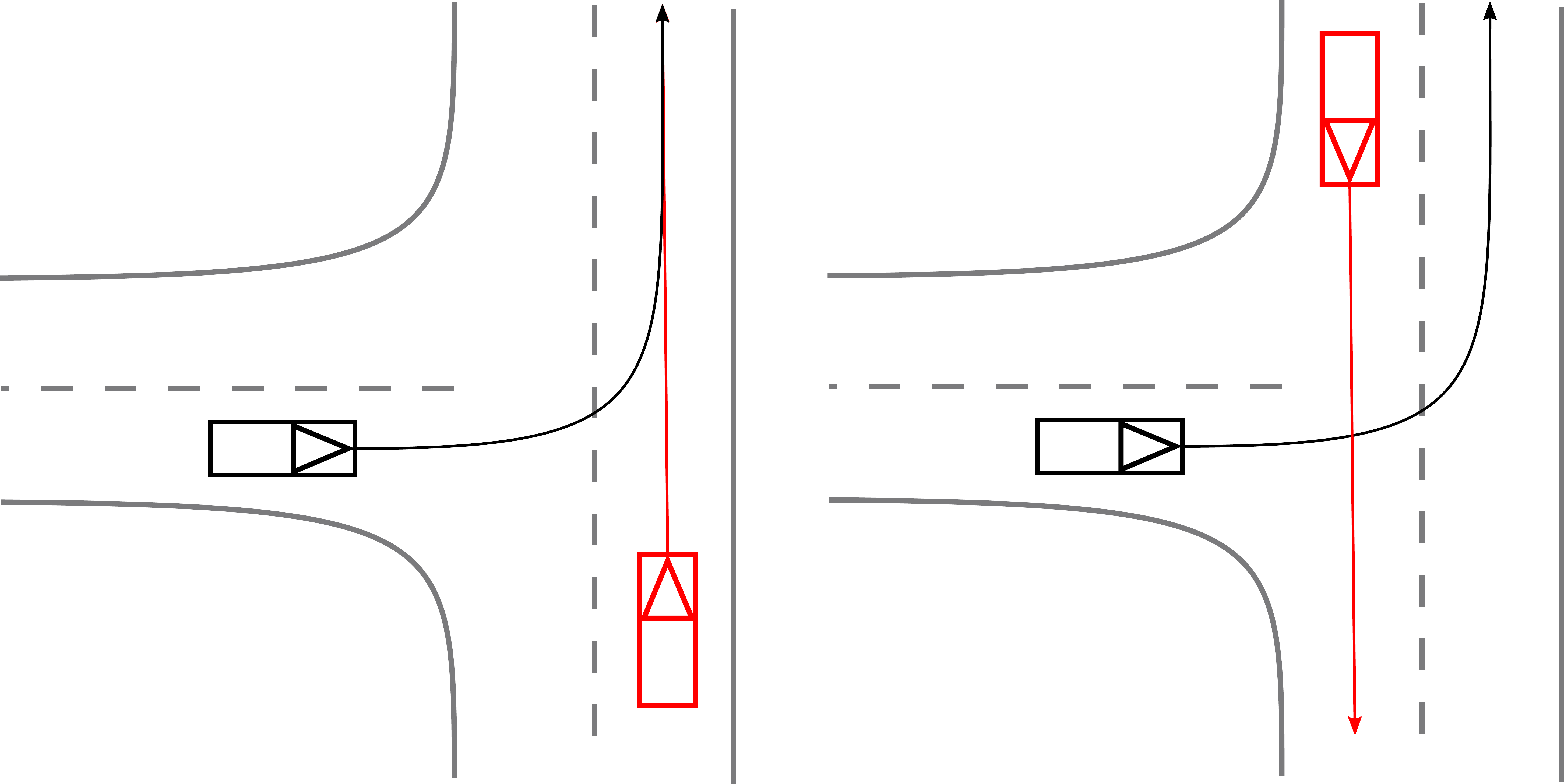}}%
    \put(0.01045719,0.46416713){\color[rgb]{0,0,0}\makebox(0,0)[lt]{\lineheight{1.25}\smash{\begin{tabular}[t]{l}1) Merging \end{tabular}}}}%
    \put(0.54177955,0.46416713){\color[rgb]{0,0,0}\makebox(0,0)[lt]{\lineheight{1.25}\smash{\begin{tabular}[t]{l}2) Crossing \end{tabular}}}}%
  \end{picture}%
\endgroup%

	\caption{Exemplary crossing and merging scenario, where the ego vehicle (black) performs a left turn without right of way. The other vehicle (red) has right of way.}
	\label{fig:crossingMerging}
\end{figure}

\subsubsection{Merging Scenario}
 A merging scenario means that the ego vehicle's and the other vehicle's center lines are merging to the same center line.
 If the ego vehicle pulls out in front of the other vehicle, costs for the other vehicle are induced. This is because the other vehicle might have to decelerate in order to keep a safety gap to the ego vehicle. Therefore, $a_{\text{inter},k}^\text{o}$ is modeled with Equation \ref{eqn:IDM} by using the ego vehicle as leader. The other vehicle is assumed to react to the ego vehicle as soon as the ego vehicle enters the intersection. 
 In contrast, $a_{\text{norm},k}^\text{o}$ is calculated without the ego vehicle as if the ego vehicle stops at the intersection.
 
\subsubsection{Crossing Scenario}
This scenario covers situations in which the ego vehicle crosses the lane of another vehicle. Therefore, a spatio-temporal constraint $\mathbf{c}^\text{e}$ is derived in which an intersection zone with a certain safety distance is not allowed to be occupied by the ego vehicle. Using these constraints, no interaction is assumed, i.e., the other vehicle does not have to react to the ego vehicle as the ego vehicle leaves the intersection zone early enough or does not enter it before the other vehicle crosses the intersection. Following this argumentation, $a_{\text{norm}}^\text{o} = a_{\text{inter}}^\text{o}$.
Thus, $j_{\text{inter}} = \infty$ if any constraint is violated. This also applies for constraints which do not emerge from crossing scenarios.

In summary, courtesy and interactions are already considered during the actual behavior planning, instead of after the trajectory generation as in \cite{Evestedt2016}.
In order to solve the presented problem, the A* graph search algorithm is employed \cite{Hubmann2016,Russell2016,Bhattachary2017}. To reduce the computational effort a heuristic function can be utilized. As presented in previous work, a reasonable approach is to exploit inevitable collision states \cite{Hubmann2016}. In the context of the presented concept this means states which definitely lead to a violation of a spatio-temporal constraint $\mathbf{c}^\text{e}$ own heuristic costs equal to the actual costs of this violation. For more details refer to \cite{Hubmann2016}.
As a result of the graph search, $\mathcal{T}_\text{B}$ is obtained and can be used for further optimization.

\subsection{Trajectory Generation}
\label{subsec:trajectoryGeneration}
\begin{figure}%
	\begin{subfigure}[c]{\textwidth}
%
%
\begin{tikzpicture}[scale=0.85]

\begin{axis}[%
width=3.2in,
height=1.19in,
at={(0.758in,0.481in)},
scale only axis,
xmin=0,
xmax=1,
ymin=-1.8,
ymax=0,
axis background/.style={fill=white},
axis x line*=bottom,
axis y line*=left,
legend style={font=\large},
legend style={legend cell align=left, align=left, draw=white!15!black},
ylabel={acceleration [$\text{m}/\text{s}^2$]}
]
\addplot [color=blue, line width=1.0pt]
  table[row sep=crcr]{%
0	0\\
0.01	-0.0058022416000001\\
0.02	-0.0224357312000004\\
0.03	-0.0487802088000009\\
0.04	-0.0837673984000015\\
0.05	-0.126380000000002\\
0.06	-0.175650681600003\\
0.07	-0.230661071200004\\
0.08	-0.290540748800005\\
0.09	-0.354466238400006\\
0.1	-0.421660000000007\\
0.11	-0.491389421600008\\
0.12	-0.562965811200009\\
0.13	-0.63574338880001\\
0.14	-0.709118278400011\\
0.15	-0.782527500000012\\
0.16	-0.855447961600012\\
0.17	-0.927395451200013\\
0.18	-0.997923628800014\\
0.19	-1.06662301840001\\
0.2	-1.13312000000002\\
0.21	-1.19707580160002\\
0.22	-1.25818549120002\\
0.23	-1.31617696880002\\
0.24	-1.37080995840002\\
0.25	-1.42187500000002\\
0.26	-1.46919244160002\\
0.27	-1.51261143120002\\
0.28	-1.55200890880002\\
0.29	-1.58728859840002\\
0.3	-1.61838000000002\\
0.31	-1.64523738160001\\
0.32	-1.66783877120001\\
0.33	-1.68618494880001\\
0.34	-1.70029843840001\\
0.35	-1.71022250000001\\
0.36	-1.71602012160001\\
0.37	-1.71777301120001\\
0.38	-1.71558058880001\\
0.39	-1.70955897840001\\
0.4	-1.69984000000001\\
0.41	-1.68657016160001\\
0.42	-1.6699096512\\
0.43	-1.6500313288\\
0.44	-1.6271197184\\
0.45	-1.60137\\
0.46	-1.5729870016\\
0.47	-1.5421841912\\
0.48	-1.5091826688\\
0.49	-1.4742101584\\
0.5	-1.43749999999999\\
0.51	-1.39929014159998\\
0.52	-1.35982213119999\\
0.53	-1.31934010879999\\
0.54	-1.27808979839998\\
0.55	-1.23631749999999\\
0.56	-1.19426908159999\\
0.57	-1.15218897119999\\
0.58	-1.11031914879999\\
0.59	-1.06889813839998\\
0.6	-1.02815999999999\\
0.61	-0.988333321599978\\
0.62	-0.949640211199984\\
0.63	-0.912295288799982\\
0.64	-0.876504678399965\\
0.65	-0.842464999999991\\
0.66	-0.81036236159998\\
0.67	-0.78037135119998\\
0.68	-0.752654028799988\\
0.69	-0.72735891839998\\
0.7	-0.704619999999985\\
0.71	-0.684555701599974\\
0.72	-0.667267891199969\\
0.73	-0.652840868799973\\
0.74	-0.641340358399983\\
0.75	-0.632812499999982\\
0.76	-0.627282841599975\\
0.77	-0.624755331199994\\
0.78	-0.625211308799998\\
0.79	-0.628608498399993\\
0.8	-0.634879999999984\\
0.81	-0.643933281599997\\
0.82	-0.655649171200013\\
0.83	-0.669880848799979\\
0.84	-0.686452838400004\\
0.85	-0.705159999999993\\
0.86	-0.725766521599989\\
0.87	-0.748004911200007\\
0.88	-0.771574988799994\\
0.89	-0.796142878400028\\
0.9	-0.82134000000001\\
0.91	-0.846762061600013\\
0.92	-0.871968051200025\\
0.93	-0.896479228800012\\
0.94	-0.919778118400019\\
0.95	-0.941307500000005\\
0.96	-0.96046940160004\\
0.97	-0.976624091200028\\
0.98	-0.989089068800071\\
0.99	-0.997138058400047\\
1	-1.00000000000002\\
};
\addlegendentry{\small{CA}}

\addplot [color=red, line width=1.0pt]
  table[row sep=crcr]{%
0	0\\
0.01	-0.000970348600000029\\
0.02	-0.00376555520000011\\
0.03	-0.00821800980000024\\
0.04	-0.0141681664000004\\
0.05	-0.0214643750000006\\
0.06	-0.0299627136000008\\
0.07	-0.0395268202000011\\
0.08	-0.0500277248000013\\
0.09	-0.0613436814000016\\
0.1	-0.0733600000000019\\
0.11	-0.0859688786000022\\
0.12	-0.0990692352000025\\
0.13	-0.112566539800003\\
0.14	-0.126372646400003\\
0.15	-0.140405625000003\\
0.16	-0.154589593600003\\
0.17	-0.168854550200004\\
0.18	-0.183136204800004\\
0.19	-0.197375811400004\\
0.2	-0.211520000000004\\
0.21	-0.225520608600004\\
0.22	-0.239334515200004\\
0.23	-0.252923469800004\\
0.24	-0.266253926400004\\
0.25	-0.279296875000004\\
0.26	-0.292027673600004\\
0.27	-0.304425880200004\\
0.28	-0.316475084800004\\
0.29	-0.328162741400004\\
0.3	-0.339480000000004\\
0.31	-0.350421538600003\\
0.32	-0.360985395200003\\
0.33	-0.371172799800003\\
0.34	-0.380988006400003\\
0.35	-0.390438125000002\\
0.36	-0.399532953600002\\
0.37	-0.408284810200001\\
0.38	-0.4167083648\\
0.39	-0.4248204714\\
0.4	-0.43264\\
0.41	-0.440187668599999\\
0.42	-0.447485875199998\\
0.43	-0.454558529799998\\
0.44	-0.461430886399997\\
0.45	-0.468129374999997\\
0.46	-0.474681433599996\\
0.47	-0.481115340199995\\
0.48	-0.487460044799995\\
0.49	-0.493745001399994\\
0.5	-0.499999999999993\\
0.51	-0.506254998599991\\
0.52	-0.512539955199992\\
0.53	-0.518884659799991\\
0.54	-0.525318566399989\\
0.55	-0.531870624999989\\
0.56	-0.538569113599988\\
0.57	-0.545441470199988\\
0.58	-0.552514124799987\\
0.59	-0.559812331399985\\
0.6	-0.567359999999987\\
0.61	-0.575179528599984\\
0.62	-0.583291635199984\\
0.63	-0.591715189799983\\
0.64	-0.60046704639998\\
0.65	-0.609561874999983\\
0.66	-0.619011993599981\\
0.67	-0.62882720019998\\
0.68	-0.639014604799981\\
0.69	-0.649578461399979\\
0.7	-0.660519999999978\\
0.71	-0.671837258599975\\
0.72	-0.683524915199975\\
0.73	-0.695574119799974\\
0.74	-0.707972326399975\\
0.75	-0.720703124999974\\
0.76	-0.73374607359997\\
0.77	-0.747076530199972\\
0.78	-0.760665484799972\\
0.79	-0.77447939139997\\
0.8	-0.78847999999997\\
0.81	-0.802624188599971\\
0.82	-0.816863795199971\\
0.83	-0.831145449799966\\
0.84	-0.845410406399967\\
0.85	-0.859594374999963\\
0.86	-0.873627353599965\\
0.87	-0.887433460199964\\
0.88	-0.90093076479996\\
0.89	-0.914031121399962\\
0.9	-0.926639999999963\\
0.91	-0.938656318599961\\
0.92	-0.949972275199961\\
0.93	-0.960473179799958\\
0.94	-0.970037286399956\\
0.95	-0.978535624999953\\
0.96	-0.985831833599956\\
0.97	-0.991781990199952\\
0.98	-0.996234444799956\\
0.99	-0.999029651399947\\
1	-0.999999999999948\\
};
\addlegendentry{\small{CJ}}
\\
\addplot [color=blue, draw=none, mark=x, mark options={solid, red}, forget plot, line width=1.0pt, mark size=4pt]
table[row sep=crcr]{%
	0	0\\
	1	-1\\
};

  table[row sep=crcr]{%
0	0\\
1	-1\\
};
\end{axis}
\end{tikzpicture}%
	\end{subfigure}
	\begin{subfigure}[c]{\textwidth}
%
%
\begin{tikzpicture}[scale=0.85]

\begin{axis}[%
width=3.2in,
height=1.19in,
at={(0.758in,0.481in)},
scale only axis,
xmin=0,
xmax=1,
ymin=0.8,
ymax=2,
axis background/.style={fill=white},
axis x line*=bottom,
axis y line*=left,
xlabel={time [$\text{s}$]},
ylabel={velocity [$\text{m}/\text{s}$]}
] 
\addplot [color=blue, line width=1.0pt]
  table[row sep=crcr]{%
0	2\\
0.01	1.999980495514\\
0.02	1.999847856896\\
0.03	1.999499416706\\
0.04	1.998843449344\\
0.05	1.99779865625\\
0.06	1.996293661184\\
0.07	1.994266515586\\
0.08	1.991664214016\\
0.09	1.988442219674\\
0.1	1.984564\\
0.11	1.980000572354\\
0.12	1.974730059776\\
0.13	1.968737256826\\
0.14	1.962013205504\\
0.15	1.95455478125\\
0.16	1.946364289024\\
0.17	1.937449069466\\
0.18	1.927821115136\\
0.19	1.917496696834\\
0.2	1.906496\\
0.21	1.894842771194\\
0.22	1.882563974656\\
0.23	1.869689458946\\
0.24	1.856251633664\\
0.25	1.84228515625\\
0.26	1.827826628864\\
0.27	1.812914305346\\
0.28	1.797587808256\\
0.29	1.781887855994\\
0.3	1.765856\\
0.31	1.749534372034\\
0.32	1.732965441536\\
0.33	1.716191783066\\
0.34	1.699255853824\\
0.35	1.68219978125\\
0.36	1.665065160704\\
0.37	1.647892863226\\
0.38	1.630722853376\\
0.39	1.613594017154\\
0.4	1.596544\\
0.41	1.579609054874\\
0.42	1.562823900416\\
0.43	1.546221589186\\
0.44	1.529833385984\\
0.45	1.51368865625\\
0.46	1.497814764544\\
0.47	1.482236983106\\
0.48	1.466978410496\\
0.49	1.452059900314\\
0.5	1.4375\\
0.51	1.423314899714\\
0.52	1.409518391296\\
0.53	1.396121837306\\
0.54	1.383134150144\\
0.55	1.37056178125\\
0.56	1.358408720384\\
0.57	1.346676504986\\
0.58	1.335364239616\\
0.59	1.324468625474\\
0.6	1.313984\\
0.61	1.303902386554\\
0.62	1.294213554176\\
0.63	1.284905087426\\
0.64	1.275962466304\\
0.65	1.26736915625\\
0.66	1.259106708224\\
0.67	1.251154868866\\
0.68	1.243491700736\\
0.69	1.236093712634\\
0.7	1.228936\\
0.71	1.221992395394\\
0.72	1.215235629056\\
0.73	1.208637499546\\
0.74	1.202169054464\\
0.75	1.19580078125\\
0.76	1.189502808064\\
0.77	1.183245114746\\
0.78	1.176997753856\\
0.79	1.170731081794\\
0.8	1.164416\\
0.81	1.158024206234\\
0.82	1.151528455936\\
0.83	1.144902833666\\
0.84	1.138123034624\\
0.85	1.13116665625\\
0.86	1.124013499904\\
0.87	1.116645882626\\
0.88	1.10904895897601\\
0.89	1.101211052954\\
0.9	1.093124\\
0.91	1.084783499074\\
0.92	1.076189474816\\
0.93	1.06734644978599\\
0.94	1.058263926784\\
0.95	1.04895678125\\
0.96	1.039445663744\\
0.97	1.029757412506\\
0.98	1.019925476096\\
0.99	1.00999034611399\\
1	0.999999999999999\\
};

\addplot [color=red, line width=1.0pt]
  table[row sep=crcr]{%
0	2\\
0.01	1.999996740969\\
0.02	1.99997451108267\\
0.03	1.999915906601\\
0.04	1.999805159424\\
0.05	1.99962805729167\\
0.06	1.999371865664\\
0.07	1.999025251281\\
0.08	1.99857820740267\\
0.09	1.998021980729\\
0.1	1.997349\\
0.11	1.99655280627567\\
0.12	1.995627984896\\
0.13	1.994570099121\\
0.14	1.99337562545067\\
0.15	1.992041890625\\
0.16	1.990567010304\\
0.17	1.98894982942767\\
0.18	1.987189864256\\
0.19	1.985287246089\\
0.2	1.98324266666667\\
0.21	1.981057325249\\
0.22	1.978732877376\\
0.23	1.97627138530767\\
0.24	1.973675270144\\
0.25	1.970947265625\\
0.26	1.96809037361067\\
0.27	1.965107821241\\
0.28	1.962003019776\\
0.29	1.95877952511567\\
0.3	1.955441\\
0.31	1.951991177889\\
0.32	1.94843382852267\\
0.33	1.944772725161\\
0.34	1.941011613504\\
0.35	1.93715418229167\\
0.36	1.933204035584\\
0.37	1.929164666721\\
0.38	1.92503943396267\\
0.39	1.920831537809\\
0.4	1.916544\\
0.41	1.91217964419567\\
0.42	1.907741078336\\
0.43	1.903230678681\\
0.44	1.89865057553067\\
0.45	1.894002640625\\
0.46	1.889288476224\\
0.47	1.88450940586767\\
0.48	1.879666466816\\
0.49	1.874760404169\\
0.5	1.86979166666667\\
0.51	1.864760404169\\
0.52	1.859666466816\\
0.53	1.85450940586767\\
0.54	1.849288476224\\
0.55	1.844002640625\\
0.56	1.83865057553067\\
0.57	1.833230678681\\
0.58	1.827741078336\\
0.59	1.82217964419567\\
0.6	1.816544\\
0.61	1.810831537809\\
0.62	1.80503943396267\\
0.63	1.799164666721\\
0.64	1.793204035584\\
0.65	1.78715418229167\\
0.66	1.781011613504\\
0.67	1.774772725161\\
0.68	1.76843382852267\\
0.69	1.761991177889\\
0.7	1.755441\\
0.71	1.74877952511567\\
0.72	1.742003019776\\
0.73	1.735107821241\\
0.74	1.72809037361067\\
0.75	1.720947265625\\
0.76	1.713675270144\\
0.77	1.70627138530767\\
0.78	1.698732877376\\
0.79	1.69105732524901\\
0.8	1.68324266666667\\
0.81	1.67528724608901\\
0.82	1.66718986425601\\
0.83	1.65894982942767\\
0.84	1.65056701030401\\
0.85	1.64204189062501\\
0.86	1.63337562545067\\
0.87	1.62457009912101\\
0.88	1.61562798489601\\
0.89	1.60655280627567\\
0.9	1.59734900000001\\
0.91	1.58802198072901\\
0.92	1.57857820740268\\
0.93	1.56902525128101\\
0.94	1.55937186566401\\
0.95	1.54962805729168\\
0.96	1.53980515942401\\
0.97	1.52991590660101\\
0.98	1.51997451108268\\
0.99	1.50999674096901\\
1	1.50000000000001\\
};

\addplot [color=blue, draw=none, mark=x, mark options={solid, blue}, forget plot, line width=1.0pt, mark size=4pt]
  table[row sep=crcr]{%
0	2\\
1	1\\
};
\addplot [color=red, draw=none, mark=x, mark options={solid, red}, forget plot,line width=1.0pt, mark size=4pt]
  table[row sep=crcr]{%
0	2\\
1	1.5\\
};
\end{axis}
\end{tikzpicture}%
	\end{subfigure}
	\caption{Exemplary comparison of directly interpolated states with the constant acceleration (CA) transition model used in the state of the art \cite{Hubmann2016} and the presented constant jerk (CJ) approach. Interpolation is done with septic polynomials, where the start state is defined by $x_0 = [\SI{0}{\meter},\SI{2}{\meter\per\second},\SI{0}{\meter\per\second\squared}]^\text{T}$ and the action by $a_0=\SI{-1}{\meter\per\second\squared}$. 
		With CA this leads to a target state $x_1 = [\SI{1.5}{\meter},\SI{1}{\meter\per\second},\SI{-1}{\meter\per\second\squared}]^\text{T}$. Using CJ results in the target state $x_1 = [\SI{1.833}{\meter},\SI{1.5}{\meter\per\second},\SI{-1}{\meter\per\second\squared}]^\text{T}$. For comparability $\dot{a}_0 =\SI{0}{\meter\per\second\cubed}$ and $\dot{a}_1 =\SI{0}{\meter\per\second\cubed}$ is set for both scenarios.}
	\label{fig:compareCaCj}
\end{figure}
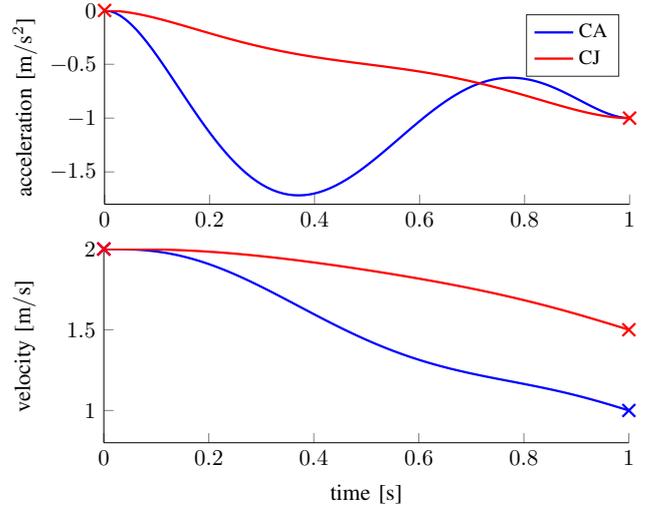

The rough discretized behavior trajectory $\mathcal{T}_\text{B}$ is used to generate a smooth and continuous execution trajectory $\mathcal{T}_\text{ex}$. In contrast to \cite{Villagra2012} and \cite{Liu2002}, the trajectories are represented by piecewise septic polynomials with fixed temporal spacing and multiple target states for different times are regarded.
In addition to acceleration continuity, septic polynomials allow for continuous jerk \cite{Rathgeber2015}. Therefore, the trajectory fulfills
\begin{equation}
\mathcal{T}_{\text{ex},i}^{(n)}(t_{i+1}) \mbeq \mathcal{T}_{\text{ex},i+1}^{(n)}(t_{i+1}) \quad\text{and}\quad n \in [0,3]\,,
\end{equation}
where $\mathcal{T}_{\text{ex},i}^{(n)}$ is the $n^{\text{th}}$ derivation of the polynomial which defines $\mathcal{T}_{\text{ex}}$ from $t_{i}$ to $t_{i+1}$.
In general, septic polynomials are defined by their coefficients $c_{0...7}$ with
\begin{equation}
\setlength\arraycolsep{1.5pt}
	\mathbf{x}(t)= \underbrace{\begin{bmatrix}
	1 & t & t^2 & t^3 \\
	0 & 1 & 2t  & 3t^2 \\
	0 & 0 & 2   & 6t \\
	0 & 0 & 0   & 6
	\end{bmatrix}}_{\mathbf{M}_1(t)}\mathbf{c}_{0123} + \setlength\arraycolsep{1.5pt}\underbrace{\begin{bmatrix}
	t^4 & t^5 & t^6 & t^7 \\
	4t^3 & 5t^4 & 6t^5  & 7t^6 \\
	12t^2 & 20t^3 & 30t^4   & 42t^5 \\
	24t & 60t^2 & 120t^3   & 210t^4
	\end{bmatrix}}_{\mathbf{M}_2(t)}\mathbf{c}_{4567}\,,
\end{equation}
where $\mathbf{x}(t)=[s,v,a,\dot{a}]^\text{T}$. Furthermore, the coefficients can be derived from a given start state $\mathbf{x}_0$, terminal state $\mathbf{x_\text{f}}$ and an according time difference $t_\text{f}$. As a result, $c_{0...7}$ can be determined by
\begin{eqnarray}
	\mathbf{c}_{0123} &=& \mathbf{M}_1^{-1}(0)\mathbf{x}_0\,,\\
	\mathbf{c}_{4567} &=& \mathbf{M}_2^{-1}(t_\text{f})(\mathbf{x}(t_\text{f}) - \mathbf{M}_1(t_\text{f})\mathbf{c}_{0123})\,.
\end{eqnarray}
For this reason, septic polynomials can be used to interpolate behavior states $\mathbf{x}_k^\text{e}$ and $\mathbf{x}_{k+1}^\text{e}$. In contrast to \cite{Hubmann2016} the utilized transition model described with Equation \ref{eqn:caTransition} enables smooth interpolations as shown in Fig. \ref{fig:compareCaCj}.
However, there might be more comfortable solutions so that not only immediately succeeding states are interpolated. With Algorithm \ref{alg:samplingTrajs} multiple trajectory candidates are generated. In general, the start state $\mathbf{x}_s^\text{e}$ is interpolated to each following state $\mathbf{x}_k^\text{e}$. The rest of the states from $\mathbf{x}_k^\text{e}$ to $\mathbf{x}_\tau^\text{e}$ can be directly interpolated.
More sophisticated solutions can be found by using the algorithm recursively in a way that each state $\mathbf{x}_k^\text{e}$ is again taken as start state. However, with the number of recursions $l_\text{r}$ the number of trajectory candidates which have to be evaluated increases. In Fig. \ref{fig:samplingExample}, some exemplary trajectory candidates are shown with according states of $\mathcal{T}_\text{B}$.

\begin{figure}
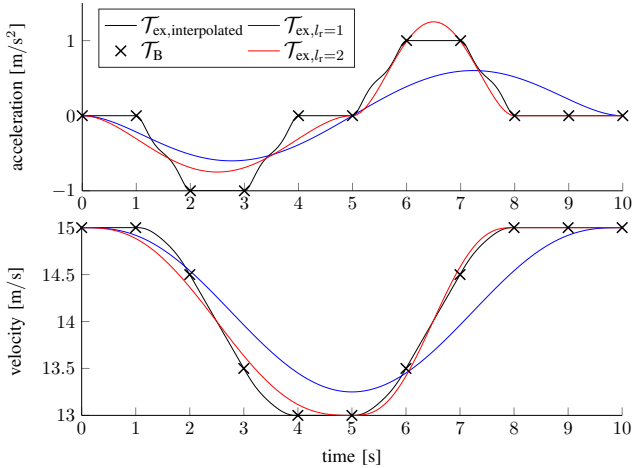

	\begin{subfigure}[c]{\textwidth}
		\input{tikzPlots/sampling_example_acc.tex}
	\end{subfigure}
	\begin{subfigure}[c]{\textwidth}
		\input{tikzPlots/sampling_example_vel.tex}
	\end{subfigure}
	\caption{Exemplary trajectory candidates for given behavior states. In black all states are directly interpolated. In blue the interpolation between first and last state. In red an example for $l_\text{r} = 2$, where the polynomials reach from $k=0$ to $k=5$ to $k=8$ and the rest of the states are directly interpolated.}
	\label{fig:samplingExample}
\end{figure}

\begin{algorithm}
	\begin{algorithmic}
	\State \texttt{get start state } $\mathbf{x}_s^\text{e}$ \texttt{ from } $\mathcal{T}_\text{B}$
	\State $k \gets s+1$  
	\While{$\mathbf k \neq \tau$}
	\State \texttt{traj} $\gets$ \texttt{interpolate($\mathbf{x}_s^\text{e}$,$\mathbf{x}_k^\text{e}$)}
	\State $j \gets k$
	\While{$j \neq \tau$}
	\State \texttt{traj.append(}\texttt{interpolate($\mathbf{x}_j^\text{e}$,$\mathbf{x}_{j+1}^\text{e}$))}
	\State $j \gets j+1$
	\EndWhile
	\State $k \gets k+1$
	\State \texttt{traj\_set.add(traj)}
	\EndWhile
\end{algorithmic}
\caption{Generation of trajectory candidates}
\label{alg:samplingTrajs}
\end{algorithm}
After the generation of trajectory candidates, these can be verified against the constraints $\mathbf{C}^\text{e}$, kinematic boundaries and evaluated with a cost functional optimizing the comfort
\begin{equation}
	J_{\text{ex}} = \int_{t_0}^{t_\tau} \dot{a}_{\text{ex}}^2 dt\,.
\end{equation}
Furthermore, there are additional constraints introduced in order to omit velocity overshooting, with
\begin{equation}
	{v}^\text{e}_{\text{min}} \leq v_{\text{ex}}(t) \leq {v}^\text{e}_{\text{max}}  \quad t \in [t_0,t_\tau]\,,
\end{equation}
where ${v}^\text{e}_{\text{min}}$ and ${v}^\text{e}_{\text{max}}$ represents the minimal and maximal velocity value contained in $\mathcal{T}_\text{B}$. Finally, the best valid trajectory can be chosen.

In contrast to state of the art concepts, the presented concept considers the whole horizon $t_{\tau}$ for the low level optimization of $\mathcal{T}_\text{B}$ \cite{Hubmann2016}. Therefore, constraints can be checked for the whole horizon and consequently safety is increased. 
In addition, the sampling effort can be drastically reduced as the behavior states can be smoothly interpolated and it is not necessary to sample states which deviate from $\mathcal{T}_\text{B}$.
As a result, smooth and safe trajectories over the whole horizon are received.

\subsection{Replanning}
\label{subsec:replanning}
\begin{figure}
	\centering
	\def\svgwidth{\columnwidth}
	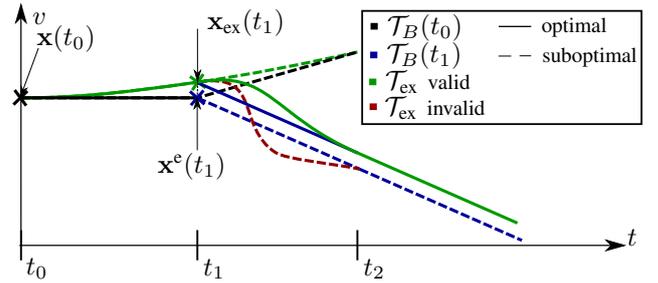
	\caption{Schematic illustration of replanning strategy with changing environment in $t_1$. The start state is denoted by $\mathbf{x}(t_0)$, the actual position in $t_1$ by $\mathbf{x}_\text{ex}(t_1)$ and the according position on the behavior trajectory by $\mathbf{x}^\text{e}(t_1)$. Dashed lines are planned states which are suboptimal and not used for driving. The behavior trajectory planned in $t_0$ is shown in black the ones planned in $t_1$ are displayed in blue. Red is the non-smooth or even invalid interpolation using the behavior trajectory starting at $\mathbf{x}^\text{e}(t_1)$. Consequently, the behavior trajectory beginning from $\mathbf{x}_\text{ex}(t_1)$ with its associated interpolation is used for the driven trajectory which is depicted in green.}
	\label{fig:replanning}
\end{figure}

As described in Section \ref{subsec:trajectoryGeneration}, $\mathcal{T}_{\text{ex}}$ might deviate from $\mathcal{T}_\text{B}$. In order to keep consistency, the start state used for behavior replanning is not the actual position $\mathbf{x}_{\text{ex}}(t)$ contained in $\mathcal{T}_{\text{ex}}$, but rather the predicted state on the behavior trajectory $\mathbf{x}^\text{e}(t)$. As a result, in static environments the behavior trajectory is consistent for subsequent planning steps. However, there is a huge impact on the solutions if the environment changes, as there might exist no smooth interpolation to the new next behavior state. For this reason, in this work an additional behavior trajectory is calculated beginning with $\mathbf{x}_\text{ex}(t)$. In this behavior trajectory a smooth interpolation of the current state to the next behavior state is guaranteed and consequently smooth reactions to changing predictions and environment. In the end, trajectory candidates for both behavior trajectories can be generated and the best candidate can be chosen. 
The according problem is illustrated in Fig. \ref{fig:replanning}.
Furthermore, as both behavior trajectories can be calculated in parallel, the additional effort should only have minor influence on the runtime.

\begin{figure*}
	 \mbox{}\hfill  
	\begin{subfigure}[b]{0.45\textwidth}
		\centering
		\includegraphics[width=\textwidth]{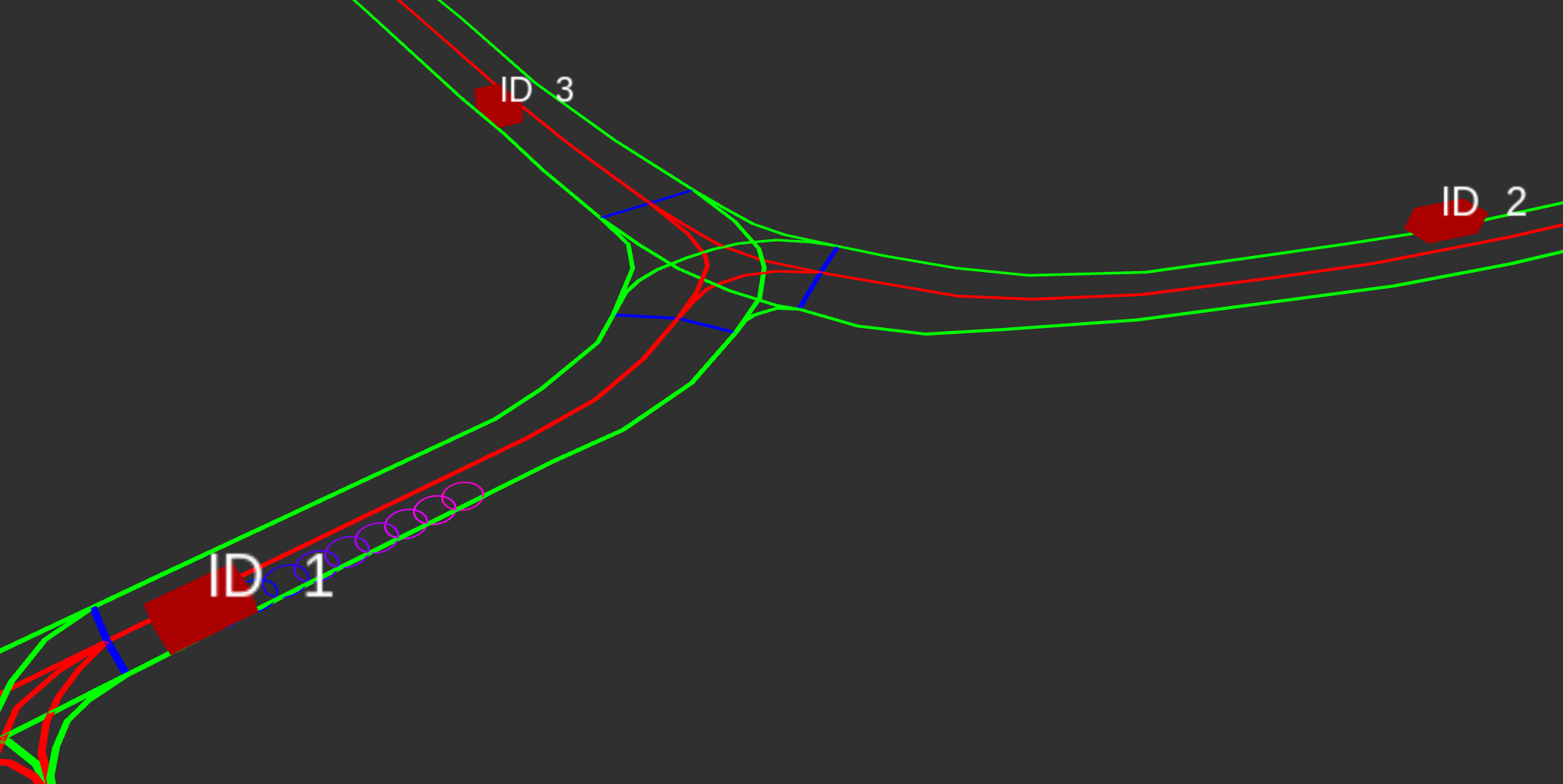}
		\caption{Exemplary overview of the scene.}
		\label{fig:evalOverview}
	\end{subfigure}
	\begin{subfigure}[b]{0.5\textwidth}
		\centering
		\input{tikzPlots/scenario_s_t_2.tex}
		\caption{Time-space diagram of driven trajectory.}
		\label{fig:evalst}
	\end{subfigure}
	   
	\begin{subfigure}[c]{0.5\textwidth}
		\centering
		\input{tikzPlots/scenario_vel_2.tex}
		\caption{Velocity of the driven trajectory.}
		\label{fig:evalv}
	\end{subfigure}
	\begin{subfigure}[c]{0.5\textwidth}
		\centering
		\input{tikzPlots/scenario_acc_2.tex}
		\caption{Acceleration of the driven trajectory.}
		\label{fig:evalacc}
	\end{subfigure}
	\begin{subfigure}[c]{0.5\textwidth}
		\centering
		\input{tikzPlots/Septic_example_scenario_vel.tex}
		\caption{Velocity for a single planned trajectory.}
		\label{fig:evalVelPlanningStep}
	\end{subfigure}
	\begin{subfigure}[c]{0.5\textwidth}
		\centering
		\input{tikzPlots/Septic_example_scenario_acc.tex}
		\caption{Acceleration for a single planned trajectory.}
		\label{fig:evalAccPlanningStep}
	\end{subfigure}
	\caption{Evaluation Scenario. (a): Exemplary overview of the scene visualized \textit{CoInCar-Sim} \cite{Naumann2018a}. Lanes are represented by lane boundaries, the ego vehicle has ID 1 and the other vehicles ID 2 and 3. %
	(b)-(d): Illustration of the driven trajectories. Scenario 1 and 2 do not include vehicle 3 and show merging behavior for different $\omega_{\text{inter}}$, where vehicle 2 is drawn in yellow. In scenario 3, vehicle 3 manifests through spatio-temporal constraint shown with a gray rectangle. %
	(e)-(f): Planned trajectory at the beginning of scenario 1, including behavior states and interpolated results.
	}
\end{figure*}

\section{Implementation and Evaluation}
For the evaluation, an intersection in Ulm-Lehr Germany is investigated and the according center lines are taken from a high-precision digital map of Ulm University \cite{Kunz2015}.
A corresponding overview is depicted in Fig. \ref{fig:evalOverview}, where the visualization is done in \textit{CoInCar-Sim} \cite{Naumann2018a}. The ego vehicle (ID 1) approaches the intersection from the left and has to perform a left turn. However, vehicle 2 (ID 2) has right of way and is approaching from the right. In one scenario, vehicle 3 (ID 3) is regarded which is approaching from top with right of way and crosses the line of the ego vehicle by driving straight.
In general, three different scenarios are regarded. In scenario 1 only the ego vehicle and vehicle 2 are involved. The ego vehicle is parametrized with $\omega^\text{l}_{\text{inter}} = 20$ which means that induced costs for other vehicles are less penalized. As a result, more aggressive behavior is expected. Scenario 2 is equal to scenario 1 except that $\omega^\text{h}_{\text{inter}} = 50$ is used. These high interaction costs will lead to more courteous behavior. In scenario 3 all vehicles are regarded and $\omega^\text{l}_{\text{inter}}$ is used. Thereby, vehicle 3 crosses the ego route before vehicle 2 passes the intersection. An overview of the scenarios with the associated interaction cost weighting and involved vehicles can be found in \mbox{Table \ref{tab:scenarios}}.
In order to emphasize the practicability of the approach and verify the robustness against uncertainties, the other vehicles are simulated with CV in contrast to the prediction model used. This means the other vehicles do not choose a decelerating reaction but rather drive aggressive. However, the ego vehicle is expected to enforce a courteous maneuver and keep an appropriate time gap to the other vehicles due to continuous replanning. 
The concept is implemented in C++ using the A*-implementation of the Discrete Optimal Search Library (DOSL) \cite{Bhattachary2017}. The according runtime is evaluated on a i5-7300U CPU with 2.6 GHz. For behavior planning the action set $\mathcal{A} = [-2,-1,0,1,2]\SI{}{\meter\per\second\squared}$ is chosen, where the maximum acceleration difference between two subsequent behavior states is set to $\Delta a = \SI{1.9}{\meter\per\square\second}$ in order to restrict the longitudinal jerk. Replanning is done with a frequency of $5\si{\hertz}$. Further parameters are presented in \mbox{Table \ref{tab:parameters}}.
{\renewcommand{\arraystretch}{1.0} 
	\begin{table}[h]
		\caption{Scenario Overview. It is shown which vehicles are regarded and which interaction cost weighting is used.}
		\label{tab:scenarios}
		{\sisetup{per-mode=fraction}
			\begin{center}
				\begin{tabular}{c| c c c c }
					Scenario&$\omega^\text{l}_{\text{inter}}$&$\omega^\text{h}_{\text{inter}}$&Vehicle 2& Vehicle 3\\
					\hline
					Scenario 1& \cmark & \xmark &\cmark&\xmark\\
					Scenario 2& \xmark & \cmark &\cmark&\xmark\\
					Scenario 3& \cmark & \xmark &\cmark&\cmark\\
					\hline
				\end{tabular}
			\end{center}
		}
	\end{table}
}

 {\renewcommand{\arraystretch}{1.0} 
 	\begin{table}[h]
 		\caption{Parameters used for evaluation}
 		\label{tab:parameters}
 		\label{table_example}
 		{\sisetup{per-mode=fraction}
 			\begin{center}
 				\begin{tabular}{c c c c c c c c}
 					\hline
 					$\omega^\text{l}_{\text{inter}}$ & 20&&$t_\tau$ & $\SI{10}{\second}$&&$T$&$\SI{1.5}{\second}$\\
 					$\omega^\text{h}_{\text{inter}}$ & 50 &&$a_{\text{max}}$& $\SI{2.5}{\meter\per\second\squared}$&&$\delta$&4\\
 					$\omega_{\text{f}}$ & 5 &&$a_{\text{min}}$& $\SI{-2.5}{\meter\per\second\squared}$&&$v_{\text{des}}$& $\SI{7.5}{\meter\per\second}$\\
 					$\omega_{v}$ & 1 &&$v_{\text{max}}$& $\SI{10}{\meter\per\second}$&&$a_{\text{IDM}}$&$\SI{0.73}{\meter\per\second\squared}$\\
 					$\omega_{\text{jerk}}$ & 1 &&$l_\text{r}$ & 1&&$b_{\text{comf}}$&$\SI{1.67}{\meter\per\second\squared}$\\
 					&&&$\Delta t$ & $\SI{1}{\second}$ &&$s_0$&$\SI{2}{\meter}$\\
 					\hline
 				\end{tabular}
 			\end{center}
 		}
 	\end{table}
 }

First, a single planning step is considered in Fig. \ref{fig:evalVelPlanningStep} and \ref{fig:evalAccPlanningStep}. Thereby, the trajectory for the first planning step in scenario 1 is shown.
In the beginning of the horizon, the speed has to be reduced in order to limit lateral acceleration due to high curvature of the center line. Subsequently, the velocity increases drastically in order to reduce the interaction costs caused by the upcoming vehicle 2. 
By sampling quintic polynomials with manual heuristics as done in previous work, such generic behavior is hard to achieve \cite{Evestedt2016}. Because on the one hand, only one target state can be set and therefore multiple objectives are difficult to enforce, as e.g. low speed at a specific longitudinal position and fast acceleration after crossing the intersection. On the other hand, the form of a single quintic polynomial might not even allow the desired behavior, as e.g. multiple subsequent acceleration and deceleration periods.
In addition to the first planning step, the driven trajectories for different scenarios are displayed in Fig. \ref{fig:evalst}-\ref{fig:evalacc}.
In scenario 1, $\omega^\text{l}_{\text{inter}}$ is used which means the induced costs are less penalized. Consequently, the ego vehicle merges into the lane in front of vehicle 2 which has right of way. However, the behavior can still be considered courteous since there is sufficient space to vehicle 2 due to a low speed reduction and fast acceleration after the crossing.
In the second scenario, induced costs are heavily penalized with $\omega^h_{\text{inter}}$. This leads to even more courteous behavior and the ego vehicle lets the other vehicle pass the intersection first and then follows with an appropriate safety gap.
In scenario 3, the ego vehicle is again parametrized with $\omega^l_{\text{inter}}$. In contrast to scenario 1, the ego vehicle has to wait for the crossing vehicle 3. As a result, the ego vehicle can not merge in front of vehicle 2 anymore since this would induce to much interaction costs even though the lower cost weighting is used.  
These results show that courteous driving and induced costs for other vehicles can be effectively considered in the presented framework.

{\renewcommand{\arraystretch}{1.0} 
	\begin{table}[h]
		\caption{Runtime evaluation, where maximum calculation time rounded to $\SI{}{\milli\second}$ is shown and dummy vehicles are added to the sate space of the single scenarios.}
		\label{tab:runtime}
			\begin{center}
				\begin{tabular}{c| c c c c c}
					Number of vehicles &1 &2&3&4&5\\
					\hline
					Scenario 1& 8 & 8 &7&9&9\\
					Scenario 2& 17 & 19 &18&20&28\\
					Scenario 3& n/a & 10 &15&13&12\\
					\hline
				\end{tabular}
			\end{center}
	\end{table}
}

The mean calculation time for the scenarios is $\SI{2.36}{\milli\second}$ and the maximum calculation time is $\SI{17}{\milli\second}$. Further runtime evaluations shown in table \ref{tab:runtime} indicate that the size of the state space has only minor influence on the runtime. Following the argumentation in Section \ref{subsec:replanning} the runtime can even be further reduced by parallelization. 
Furthermore, smooth trajectories are generated even though the other vehicles do not behave as predicted in the framework with the IDM but rather with CV.
This additionally emphasizes that prediction inaccuracies can be overcome by continuous replanning and the robustness of the approach.

\section{Conclusion}
In this work, a novel, real time capable behavior and trajectory planning concept is presented. Induced costs for other traffic participants are considered by a game-theoretic approach and a courteous driving strategy is generated. 
The formulated optimal control problem is solved using a discrete graph-based approach including a transition model that allows for smooth interpolation between the states. Further, a trajectory generation concept based on septic polynomials is shown that enables the generation of smooth and safe trajectories on large horizons. 
Future work will include measurement and prediction uncertainties, e.g., as shown in \cite{Ward2018a}. Additionally, the concept will be extended to other traffic use cases and implemented on the experimental vehicle of Ulm University.




\bibliographystyle{../IEEEtranBST/IEEEtran.bst}

\begin{thebibliography}{10}
	\providecommand{\url}[1]{#1}
	\csname url@rmstyle\endcsname
	\providecommand{\newblock}{\relax}
	\providecommand{\bibinfo}[2]{#2}
	\providecommand\BIBentrySTDinterwordspacing{\spaceskip=0pt\relax}
	\providecommand\BIBentryALTinterwordstretchfactor{4}
	\providecommand\BIBentryALTinterwordspacing{\spaceskip=\fontdimen2\font plus
		\BIBentryALTinterwordstretchfactor\fontdimen3\font minus
		\fontdimen4\font\relax}
	\providecommand\BIBforeignlanguage[2]{{%
			\expandafter\ifx\csname l@#1\endcsname\relax
			\typeout{** WARNING: IEEEtran.bst: No hyphenation pattern has been}%
			\typeout{** loaded for the language `#1'. Using the pattern for}%
			\typeout{** the default language instead.}%
			\else
			\language=\csname l@#1\endcsname
			\fi
			#2}}
	
	\bibitem{Villagra2012}
	\BIBentryALTinterwordspacing
	J.~Villagra, V.~Milanés, J.~Pérez, and J.~Godoy, ``Smooth path and speed
	planning for an automated public transport vehicle,'' \emph{Robotics and
		Autonomous Systems}, vol.~60, no.~2, pp. 252 -- 265, 2012. [Online].
	Available:
	\url{http://www.sciencedirect.com/science/article/pii/S092188901100203X}
	\BIBentrySTDinterwordspacing
	
	\bibitem{Werling2010}
	M.~Werling, J.~Ziegler, S.~Kammel, and S.~Thrun, ``Optimal trajectory
	generation for dynamic street scenarios in a fren x00e9;t frame,'' in
	\emph{2010 IEEE International Conference on Robotics and Automation}, May
	2010, pp. 987--993.
	
	\bibitem{Ziegler2014}
	J.~Ziegler, P.~Bender, T.~Dang, and C.~Stiller, ``Trajectory planning for
	bertha 2014; a local, continuous method,'' in \emph{2014 IEEE Intelligent
		Vehicles Symposium Proceedings}, June 2014, pp. 450--457.
	
	\bibitem{Kunz2015}
	F.~Kunz, D.~Nuss, J.~Wiest, H.~Deusch, S.~Reuter, F.~Gritschneder, A.~Scheel,
	M.~Stübler, M.~Bach, P.~Hatzelmann, C.~Wild, and K.~Dietmayer, ``Autonomous
	driving at ulm university: A modular, robust, and sensor-independent fusion
	approach,'' in \emph{2015 IEEE Intelligent Vehicles Symposium (IV)}, June
	2015, pp. 666--673.
	
	\bibitem{Evestedt2016}
	N.~Evestedt, E.~Ward, J.~Folkesson, and D.~Axehill, ``Interaction aware
	trajectory planning for merge scenarios in congested traffic situations,'' in
	\emph{2016 IEEE 19th International Conference on Intelligent Transportation
		Systems (ITSC)}, Nov 2016, pp. 465--472.
	
	\bibitem{Sun2018}
	L.~Sun, W.~Zhan, M.~Tomizuka, and A.~D. Dragan, ``Courteous autonomous cars,''
	\emph{arXiv preprint arXiv:1808.02633}, 2018.
	
	\bibitem{Hubmann2016}
	C.~Hubmann, M.~Aeberhard, and C.~Stiller, ``A generic driving strategy for
	urban environments,'' in \emph{2016 IEEE 19th International Conference on
		Intelligent Transportation Systems (ITSC)}, Nov 2016, pp. 1010--1016.
	
	\bibitem{Ziegler2009}
	J.~Ziegler and C.~Stiller, ``Spatiotemporal state lattices for fast trajectory
	planning in dynamic on-road driving scenarios,'' in \emph{2009 IEEE/RSJ
		International Conference on Intelligent Robots and Systems}, Oct 2009, pp.
	1879--1884.
	
	\bibitem{Ward2018a}
	E.~Ward and J.~Folkesson, ``Towards risk minimizing trajectory planning in
	on-road scenarios,'' in \emph{2018 IEEE Intelligent Vehicles Symposium (IV)},
	June 2018, pp. 490--497.
	
	\bibitem{Brechtel2014}
	S.~Brechtel, T.~Gindele, and R.~Dillmann, ``Probabilistic decision-making under
	uncertainty for autonomous driving using continuous pomdps,'' in \emph{17th
		International IEEE Conference on Intelligent Transportation Systems (ITSC)},
	Oct 2014, pp. 392--399.
	
	\bibitem{Hubmann2018}
	C.~Hubmann, J.~Schulz, M.~Becker, D.~Althoff, and C.~Stiller, ``Automated
	driving in uncertain environments: Planning with interaction and uncertain
	maneuver prediction,'' \emph{IEEE Transactions on Intelligent Vehicles},
	vol.~3, no.~1, pp. 5--17, March 2018.
	
	\bibitem{Bai2015}
	H.~{Bai}, S.~{Cai}, N.~{Ye}, D.~{Hsu}, and W.~S. {Lee}, ``Intention-aware
	online pomdp planning for autonomous driving in a crowd,'' in \emph{2015 IEEE
		International Conference on Robotics and Automation (ICRA)}, May 2015, pp.
	454--460.
	
	\bibitem{Treiber2000}
	M.~Treiber, A.~Hennecke, and D.~Helbing, ``Congested traffic states in
	empirical observations and microscopic simulations,'' \emph{Physical review
		E}, vol.~62, no.~2, p. 1805, 2000.
	
	\bibitem{Russell2016}
	S.~J. Russell and P.~Norvig, \emph{Artificial intelligence: a modern
		approach}.\hskip 1em plus 0.5em minus 0.4em\relax Malaysia; Pearson Education
	Limited,, 2016.
	
	\bibitem{Bhattachary2017}
	\BIBentryALTinterwordspacing
	S.~Bhattacharya, ``Discrete optimal search library (dosl): A template-based c++
	library for discrete optimal search,'' 2017. [Online]. Available:
	\url{https://github.com/subh83/DOSL}
	\BIBentrySTDinterwordspacing
	
	\bibitem{Liu2002}
	S.~{Liu}, ``An on-line reference-trajectory generator for smooth motion of
	impulse-controlled industrial manipulators,'' in \emph{7th International
		Workshop on Advanced Motion Control. Proceedings (Cat. No.02TH8623)}, July
	2002, pp. 365--370.
	
	\bibitem{Rathgeber2015}
	C.~Rathgeber, F.~Winkler, X.~Kang, and S.~M{\"u}ller, ``Optimal trajectories
	for highly automated driving,'' \emph{World Academy of Science, Engineering
		and Technology, International Journal of Mechanical, Aerospace, Industrial,
		Mechatronic and Manufacturing Engineering}, vol.~9, no.~6, pp. 969--975,
	2015.
	
	\bibitem{Naumann2018a}
	M.~Naumann, F.~Poggenhans, M.~Lauer, and C.~Stiller, ``Coincar-sim: An
	open-source simulation framework for cooperatively interacting automobiles,''
	in \emph{IEEE Intl. Conf. Intelligent Vehicles}, 2018.
	
\end{thebibliography}

\end{document}